%% file: main.tex
\documentclass[runningheads]{llncs}
\usepackage{graphicx}

\usepackage{tikz}
\usepackage{comment}
\usepackage{amsmath,amssymb} 
\usepackage{color}
\usepackage{enumitem}
\usepackage{caption}
\usepackage{subcaption}
\usepackage[pagebackref,breaklinks,colorlinks,bookmarks=false]{hyperref}
\usepackage{multirow}
\usepackage{booktabs}
\usepackage{wrapfig}
\usepackage{courier}
\usepackage{orcidlink}

\newcommand{\eg}{\textit{e}.\textit{g}.}
\newcommand{\cf}{\textit{c}.\textit{f}.}

\usepackage[accsupp]{axessibility}  

\begin{document}
\pagestyle{headings}
\mainmatter
\def\ECCVSubNumber{7346}  

\newcommand{\cs}[1]{{\color{blue}[Cordelia: #1]}}

\title{Learning from Unlabeled 3D Environments for Vision-and-Language Navigation} 

\titlerunning{Learning from Unlabeled 3D Environments for VLN}
%
\author{Shizhe Chen\inst{1}\orcidlink{0000-0002-7313-9703} \and
Pierre-Louis Guhur\inst{1} \and
Makarand Tapaswi\inst{2}\orcidlink{ 0000-0001-8800-9015}
\and
\\ Cordelia Schmid\inst{1}
\and
Ivan Laptev\inst{1}\orcidlink{0000-0001-7072-3325}
}
%
\authorrunning{S. Chen et al.}
\institute{
Inria, École normale supérieure, CNRS, PSL Research University 
\and
IIIT Hyderabad
}

\maketitle

{\footnotesize\quad\quad \url{https://cshizhe.github.io/projects/hm3d_autovln.html}}

\begin{abstract}
In vision-and-language navigation (VLN), an embodied agent is required to navigate in realistic 3D environments following natural language instructions.
One major bottleneck for existing VLN approaches is the lack of sufficient training data, resulting in unsatisfactory generalization to unseen environments.
While VLN data is typically collected manually, such an approach is expensive and prevents scalability.
In this work, we address the data scarcity issue by proposing to automatically create a large-scale VLN dataset from 900 unlabeled 3D buildings from HM3D~\cite{ramakrishnan2021hm3d}.
We generate a navigation graph for each building and transfer object predictions from 2D to generate pseudo 3D object labels by cross-view consistency. 
We then fine-tune a pretrained language model using pseudo object labels as prompts to alleviate the cross-modal gap in instruction generation.
Our resulting HM3D-AutoVLN dataset is an order of magnitude larger than existing VLN datasets in terms of navigation environments and instructions. 
We experimentally demonstrate that HM3D-AutoVLN significantly increases the generalization ability of resulting VLN models. 
On the SPL metric, our approach improves over state of the art by 7.1\% and 8.1\% on the unseen validation splits of REVERIE and SOON datasets respectively.

\end{abstract}

\input{introduction}
\input{related_works}

\input{method_data}

\input{method_model}
\input{exprs}

\input{conclusion}

{\small
\noindent
\textbf{Acknowledgements.}
This work was granted access to the HPC resources of IDRIS under the allocation 101002 made by GENCI. 
This work is funded in part by the French government under management of Agence Nationale de la Recherche as part of the ``Investissements d'avenir'' program, reference ANR19-P3IA-0001 (PRAIRIE 3IA Institute) and by Louis Vuitton ENS Chair on Artificial Intelligence.
}

\input{main.bbl}
\newpage
\appendix
\include{suppmat}

\end{document}

%% file: introduction.tex
\section{Introduction}

Having a robot carry out various chores has been a common vision from science fiction. Such a long-term goal requires an embodied agent to understand our human language, navigate in the physical environment and interact with objects.
As an initial step towards this goal, the vision-and-language navigation task~(VLN)~\cite{anderson2018vision} has emerged and attracted growing research attention.
Early VLN approaches~\cite{anderson2018vision,ku2020room} provide agents with step-by-step navigation instructions to arrive at a target location, such as \emph{``Walk out of the bedroom. Turn right and walk down the hallway. At the end of the hallway turn left. Walk in front of the couch and stop"}. While these detailed instructions reduce the difficulty of the task, they lower the practical value for people when commanding robots in real life.
Thus, more recent VLN methods~\cite{qi2020reverie,zhu2021soon} focus on high-level instructions and request an agent to find a specific object at the target location, \eg,~\emph{``Go to the living room and bring me the white cushion on the sofa closest to the lamp"}. The agent needs to explore 3D environments on its own to find the \emph{``cushion''}.

\begin{figure}[t]
\centering
\includegraphics[width=\linewidth]{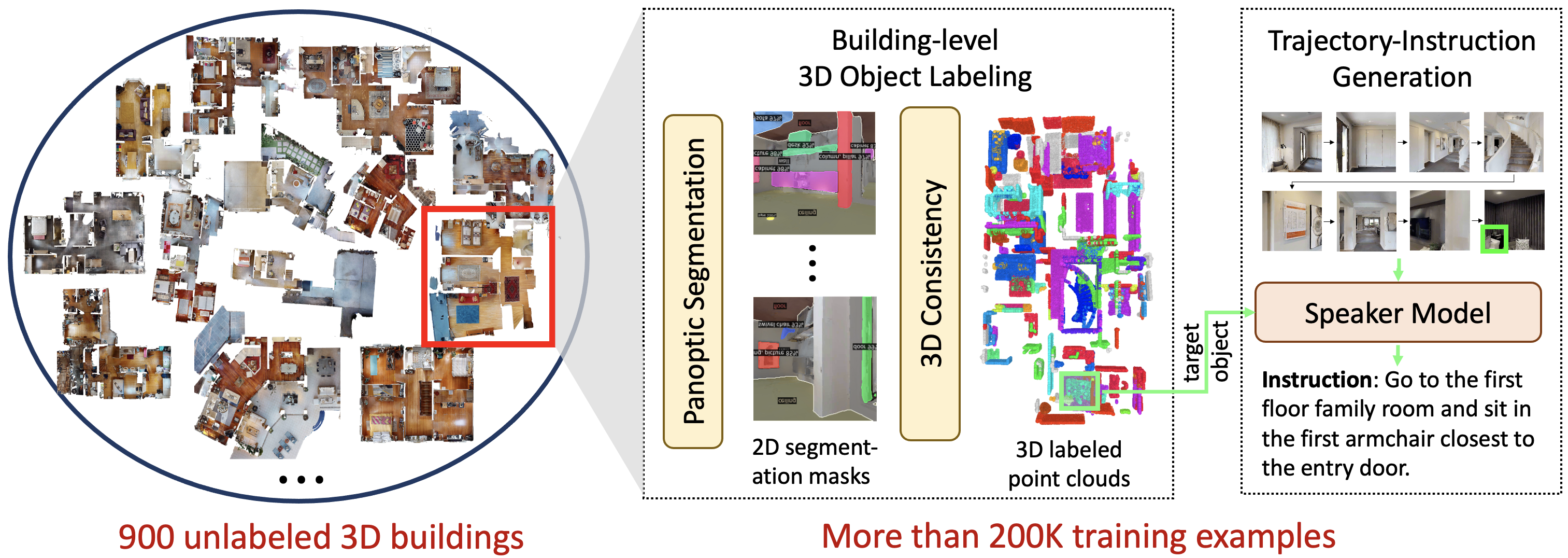}
\caption{{\bf HM3D-AutoVLN dataset}. We use 900 unlabeled 3D buildings from the HM3D~\cite{ramakrishnan2021hm3d} dataset. We improve labels obtained with a 2D segmentation model based on 3D cross-view consistency (middle, see Fig.~\ref{fig:data_generation_object}). Then we rely on these pseudo labels to generate instructions via a speaker model (right, see Fig.~\ref{fig:data_generation_speaker}). We automatically create over 200K realistic training samples for VLN.}
\label{fig:intro}
\end{figure}

Compared to step-by-step instructions, following such high-level goal-driven instructions is more challenging.
As there is no detailed guidance, an agent needs to learn
about the structure of the environment for effective exploration. 
However, most existing VLN tasks such as REVERIE~\cite{qi2020reverie} or SOON~\cite{zhu2021soon} are based on 3D scans from the Matterport3D (MP3D) dataset~\cite{chang2017matterport3d}, and contain less than 60 buildings and approximately 10K training trajectories.
The limited amount of training data makes VLN models overfit to seen environments, resulting in less generalizable navigation policies in unseen environments.
Manually collecting more VLN data is however expensive and not scalable.
To address this data scarcity issue, previous works have investigated various data augmentation methods, such as synthesizing more instructions and trajectories in seen environments by a speaker model~\cite{fried2018speaker}, environment dropout~\cite{tan2019learning} or editing~\cite{li2022envedit}, and mixing up seen environments~\cite{liu2021vision}.
Nevertheless, these approaches are still based on a small amount of 3D environments that cannot cover a wide range of objects and scenes.
To address visual diversity, VLN-BERT~\cite{majumdar2020improving} utilizes image-caption pairs from the web to improve generalization ability, while Airbert~\cite{guhur2021airbert} shows that pairs from indoor environments, the BnB dataset, are more beneficial for VLN tasks.
However, it is hard for image-caption pairs to mimic the real navigation experience of an agent in 3D environments making it challenging to learn a navigation policy with action prediction.

In this work, we propose a new data generation approach to improve the model's generalization ability to unseen environments by learning from large-scale unlabeled 3D buildings (see Fig.~\ref{fig:intro}).
We take advantage of the recent HM3D dataset~\cite{ramakrishnan2021hm3d} consisting of 900 buildings in 3D.
However, this data comes without any labels. 
In order to generate high-quality instruction-trajectory pairs on a diverse set of unseen environments, we use large-scale pretrained vision and language models.
We first use an image segmentation model~\cite{cheng2021mask2former} to detect 2D objects for images in the environment, and utilize cross-view consistency in 3D to increase the accuracy of pseudo 3D object annotations. 
Then, we fine-tune a language model, GPT-2~\cite{radford2019gpt2}, with pseudo object labels as prompts to generate high-level navigation instructions to this object.
In this way, we construct the HM3D-AutoVLN dataset that uses 900 3D buildings, and consists of 36,562 navigable nodes, 172,000 3D objects and 217,703 object-instruction-trajectory triplets for training -- an order of magnitude larger than prior VLN datasets.
We train multiple state-of-the-art VLN models~\cite{chen2021hamt,chen2022duet,hong2020recurrent,tan2019learning} with the generated HM3D-AutoVLN data and show significant gains. 
Specifically, we improve over the state-of-the-art DUET model~\cite{chen2022duet} on REVERIE and SOON datasets by 7.1\% and 8.1\% respectively.
In summary, our contributions are as follows:
\parskip=0.1em
\begin{itemize}[itemsep=0.1em,parsep=0em,topsep=0em,partopsep=0em]
	\item We introduce an automatic approach to construct a large-scale VLN dataset, HM3D-AutoVLN, from unlabeled 3D buildings. We rely on 2D image models to obtain pseudo 3D object labels and on pretrained language models to generate instructions.
	\item We carry out extensive experiments on two challenging VLN tasks, REVERIE and SOON. The training on HM3D-AutoVLN dataset significantly improves the performance for multiple state-of-the-art VLN models.
	\item We provide insights on data collection and challenges inherent to leveraging unlabeled environments. It suggests that the diversity of environments is more important than the number of training samples alone.
\end{itemize}

%% file: related_works.tex
\section{Related Work}

\subsubsection{Vision-and-language navigation.}
The VLN tasks have recently been popularized with the emergence of various supportive datasets \cite{anderson2018vision,chen2019touchdown,irshad2021robovln,krantz2020beyond,ku2020room,padmakumar2022teach,shridhar2020alfred,thomason2020vision}.
Different instructions define the variations of VLN tasks. Step-by-step instructions~\cite{anderson2018vision,chen2019touchdown,ku2020room} require an agent to strictly follow the path, whereas goal-driven high-level instructions~\cite{qi2020reverie,zhu2021soon} mainly describe the target place and object and command the agent to retrieve a particular remote object.
The embodied question answering task asks an agent to navigate and answer a question~\cite{das2018embodied,yu2019multi}, and vision-and-language dialog~\cite{de2018talk,nguyen2019hanna,padmakumar2022teach,thomason2020vision} tasks use interactive communications with agents to guide the navigation.
Due to the inherent multimodal nature, works on VLN tasks provide appealing and creative model architectures, such as 
cross-modal attention mechanism~\cite{fried2018speaker,ma2019self,ma2019regretful,tan2019learning,wang2020active},
awareness of objects~\cite{hong2020language,moudgil2021soat,qi2021road},
sequence modeling using transformers~\cite{chen2021hamt,hong2020recurrent,pashevich2021episodic},
Bayesian state tracking~\cite{anderson2019chasing},
and graph-based structures for better exploration~\cite{chen2022duet,deng2020evolving,wang2021structured}.
The models are usually pretrained in a teacher forcing manner~\cite{chen2021hamt,hao2020towards} and then fine-tuned using student forcing~\cite{anderson2018vision,chen2022duet}, stochastic sampling~\cite{li2019press}, or reinforcement learning~\cite{chen2021hamt,hong2020recurrent,tan2019learning,wang2019reinforced}.
While most existing VLN works focus on discrete environment with predefined navigation graphs, VLN in continuous environments~\cite{krantz2021waypoint,krantz2020beyond} is more practical in real world~\cite{anderson2021sim}.
The discrete environments also prove to be beneficial for continuous VLN such as providing waypoint supervision~\cite{raychaudhuri2021language} and enabling hierarchical modeling of low-level and high-level actions~\cite{chen2021topological,hong2022bridging}.

\subsubsection{Data-centric VLN approaches.}
One of the major challenges of VLN remains the scarcity of training data, leading to a large gap between seen and unseen environments.
Data augmentation is one effective approach to address over-fitting, such as to dropout environment features~\cite{tan2019learning}, change image styles~\cite{li2022envedit}, mixup rooms in different environments~\cite{liu2021vision}, generate more path-instruction pairs from speaker models~\cite{fried2018speaker,fu2020counterfactual,tan2019learning} and new images from GANs~\cite{koh2021pathdreamer}.
However, those data augmentation methods are still based on a limited number of environments, abating generalization ability on unseen environments.
Contrary to the above, VLNBert~\cite{majumdar2020improving} and Airbert~\cite{guhur2021airbert} exploit abundant web image-captions to improve generalization.
Nevertheless, such data do not fully resemble real navigation experience and thus they can only be used to train a compatibility model to measure path-instruction similarity instead of learning a navigation policy.
In our work, photo-realistic 3D environments are used to learn a navigation policy.

\subsubsection{3D environments.}
3D environments and simulation platforms~\cite{Kolve2017AI2THORAI,li2022igibson,puig2018virtualhome,savva2019habitat} are a basic foundation to promote research in embodied intelligence.
There are artificial environments based on video game engines such as iThor~\cite{Kolve2017AI2THORAI} and VirtualHome~\cite{puig2018virtualhome}, which utilize synthetic scenes and allow interactions with objects.
However, synthetic scenes have limited visual diversity and do not fully reflect the real world.
Thus, we focus on photo-realistic 3D environments for VLN.
The MP3D~\cite{chang2017matterport3d} dataset is a collection of 90 labeled environments and is most widely adopted in existing VLN datasets such as R2R~\cite{anderson2018vision}, RxR~\cite{ku2020room}, REVERIE~\cite{qi2020reverie} and SOON~\cite{zhu2021soon}. Though possessing high-quality reconstructions, the number of houses in MP3D is limited.
The Gibson~\cite{xia2018gibson} dataset contains 571 scenes, however they have poor 3D reconstruction quality.
The recent HM3D dataset~\cite{ramakrishnan2021hm3d} has the largest number of 3D environments though with no labels. It contains 1,000 high-quality and diverse building-scale reconstructions (900 publicly released) around the world such as multi-floor residences, stores, offices, \emph{etc}.
In this work, we employ the unlabeled HM3D dataset to scale up VLN datasets.

%% file: method_data.tex
\section{Generating VLN Dataset from Unlabeled 3D Buildings}

In this section, we present our approach to automatically generate large-scale VLN data from unlabeled 3D environments, specifically HM3D~\cite{ramakrishnan2021hm3d}.
We consider a discrete navigation environment~\cite{anderson2018vision,qi2020reverie,zhu2021soon}, which is the mainstream setting for VLN.
The discrete setup treats each environment as an undirected graph $\mathcal{G}=\{\mathcal{V}, \mathcal{E}\}$, where $\mathcal{V}$ denotes navigable nodes, and $\mathcal{E}$ denotes connectivity edges.
Each node $V_i \in \mathcal{V}$ corresponds to a panoramic RGB-D image captured at its location $(x, y, z)$. An agent is equipped with a camera and GPS sensors and moves between connected nodes on this navigation graph.

We aim to generate VLN data with high-level instructions (similar to~\cite{qi2020reverie,zhu2021soon}), which mainly describe the target scene and specific object instance.
The success of the agent is defined as arriving at a node where the target object is close and visible and correctly localizing the object.
Therefore, the training data should consist of a language instruction, an initial position and orientation of the agent, a ground-truth trajectory to the target location and the position of the target object.
In order to generate such training data, it is essential to know object locations in the 3D space instead of bounding boxes in 2D images. Otherwise, it can be hard to infer relationships between objects in different views leading to unreliable target locations.
Moreover, a sentence that refers to a specific object should be generated since the object name alone is not specific.
Generation of such data is challenging because the original 3D buildings come without any labels. 
Therefore, we propose to use image and text models pretrained on large-scale external datasets for pseudo labeling.
This involves two stages: (i) building-level pre-processing to obtain navigation graphs and 3D object annotations (Sec.~\ref{sec:hm3d_obj_anno}); and (ii) object-trajectory-instruction triplet generation (Sec.~\ref{sec:hm3d_vln_instr}).

\begin{figure}[t]
\centering
    \includegraphics[width=1\linewidth]{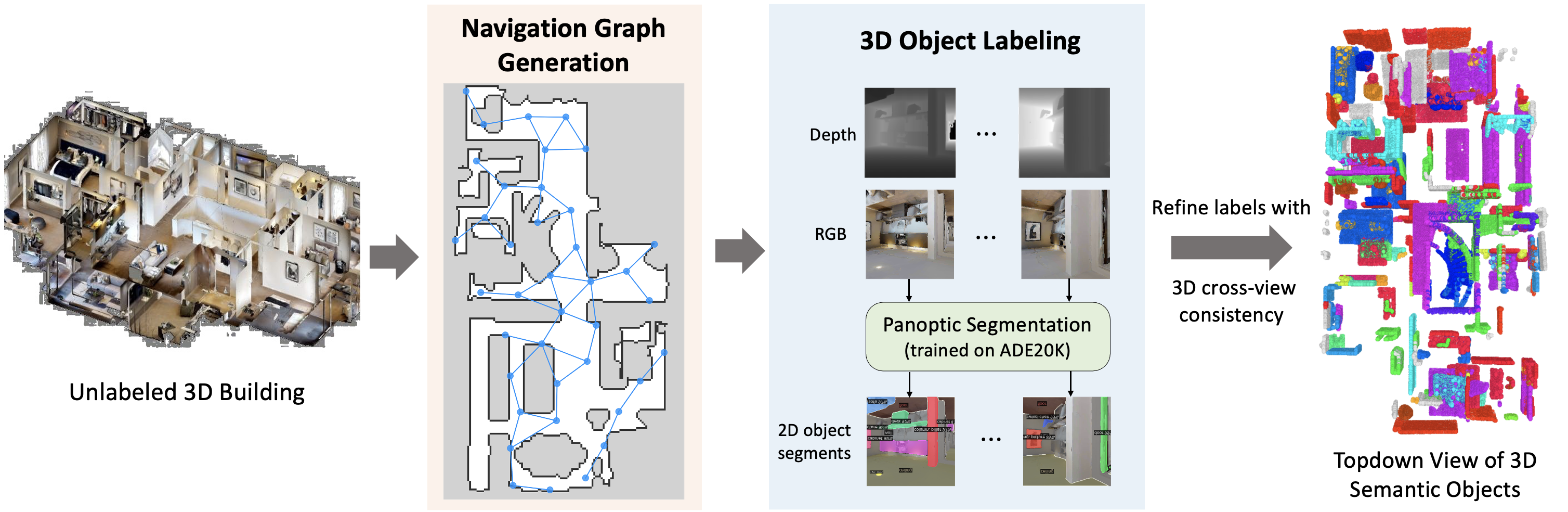}
    \caption{We automatically construct the navigation graph and label 3D objects for each unlabeled 3D building.}
    \label{fig:data_generation_object}
\end{figure}

\subsection{Building-level Pre-processing}
\label{sec:hm3d_nav_graph}

We utilize the Habitat simulator~\cite{savva2019habitat} to load 3D buildings in HM3D~\cite{ramakrishnan2021hm3d} such that agents can navigate.
This is a continuous navigation environment, thus we first convert each environment into a discrete navigation graph and then extract 3D pseudo object annotations for each building as illustrated in Fig.~\ref{fig:data_generation_object}.

\subsubsection{Constructing navigation graphs.}
We follow the characteristics of navigation graphs in the MP3D simulator~\cite{anderson2018vision} to create new graphs for continuous HM3D.
For example, the average distance between connected nodes is around 2 meters and connectivity is defined by whether two nodes are visible and navigable from each other.
However, the MP3D simulator relies on human efforts to create such graphs where the nodes are selected and navigability is inspected manually. Instead, we build the graph in a fully automatic way. 
We first randomly sample 20,000 navigable locations in the 3D environment and then greedily add locations as new nodes into the graph. The newly added location is the nearest one to existing nodes in the graph among all remaining candidates that are more than 2 meters away from the existing nodes. 
After collecting all the nodes, we use two criteria to connect different nodes: (i)~the geodesic distance between nodes is below 3 meters; and (ii) the average distance of the depth image captured from one point to another should be larger than 2 meters.
Fig.~\ref{fig:data_generation_object} (left) presents an example of navigation graph generation on one floor of the environment.
For each node, we extract 36 RGB-D images from different orientations to represent the panoramic view, following the VLN setting~\cite{anderson2018vision}.

\subsubsection{Labeling 3D objects from 2D predictions.}
\label{sec:hm3d_obj_anno}
Since existing 3D datasets such as Scannet~\cite{dai2017scannet} are small and contain a limited number of object classes, pretrained 3D object detectors are incapable of providing high-quality 3D object labels for HM3D which covers a diverse set of scenes and objects.
Therefore, we propose to use an existing 2D image model to label 3D objects.
Specifically, we use a panoptic segmentation model Mask2Former~\cite{cheng2021mask2former} trained on ADE20K~\cite{zhou2017scene} to generate instance masks for all images in the constructed graph.
We project 2D pixel-wise semantic predictions into 3D point clouds using the camera intrinsic parameters and depth information, and thus obtain 3D bounding boxes.
However, due to the domain gap between existing 2D labeled images and images from HM3D, the predictions of 2D models can be noisy as shown in Fig.~\ref{fig:quality_pseudolabels}. Even worse, a 3D object is often partially observed from one view, making the estimated 3D bounding box from a single location less accurate.
In order to further improve 3D object labeling, we take advantage of cross-view consistency of 3D objects and merge 2D predictions from multiple views.
To ease computation and reduce label noise, we downsample the extracted point clouds with class probabilities into voxels of size $0.1 \times 0.1 \times 0.1 m^3$ in a similar way as semantic map construction~\cite{chaplot2020object}. Hence, 2D predictions from different views of an object can be merged together. We average class probabilities of all points inside each voxel and take the label with maximum probability to refine semantic prediction. 
The neighboring voxels of the same class are grouped together as a complete 3D object.
In this way, we generate 3D objects for the whole building and also map 2D objects of different views into the unified 3D objects. This enables us to create reliable goal locations for target objects in the VLN task.
Fig.~\ref{fig:data_generation_object} (right) shows example predictions of 3D semantic objects from a top-down view.

\subsection{Data Generation: VLN Training Triplets}
\label{sec:hm3d_vln_instr}

\begin{figure}[t]
\centering
\includegraphics[width=0.8\linewidth]{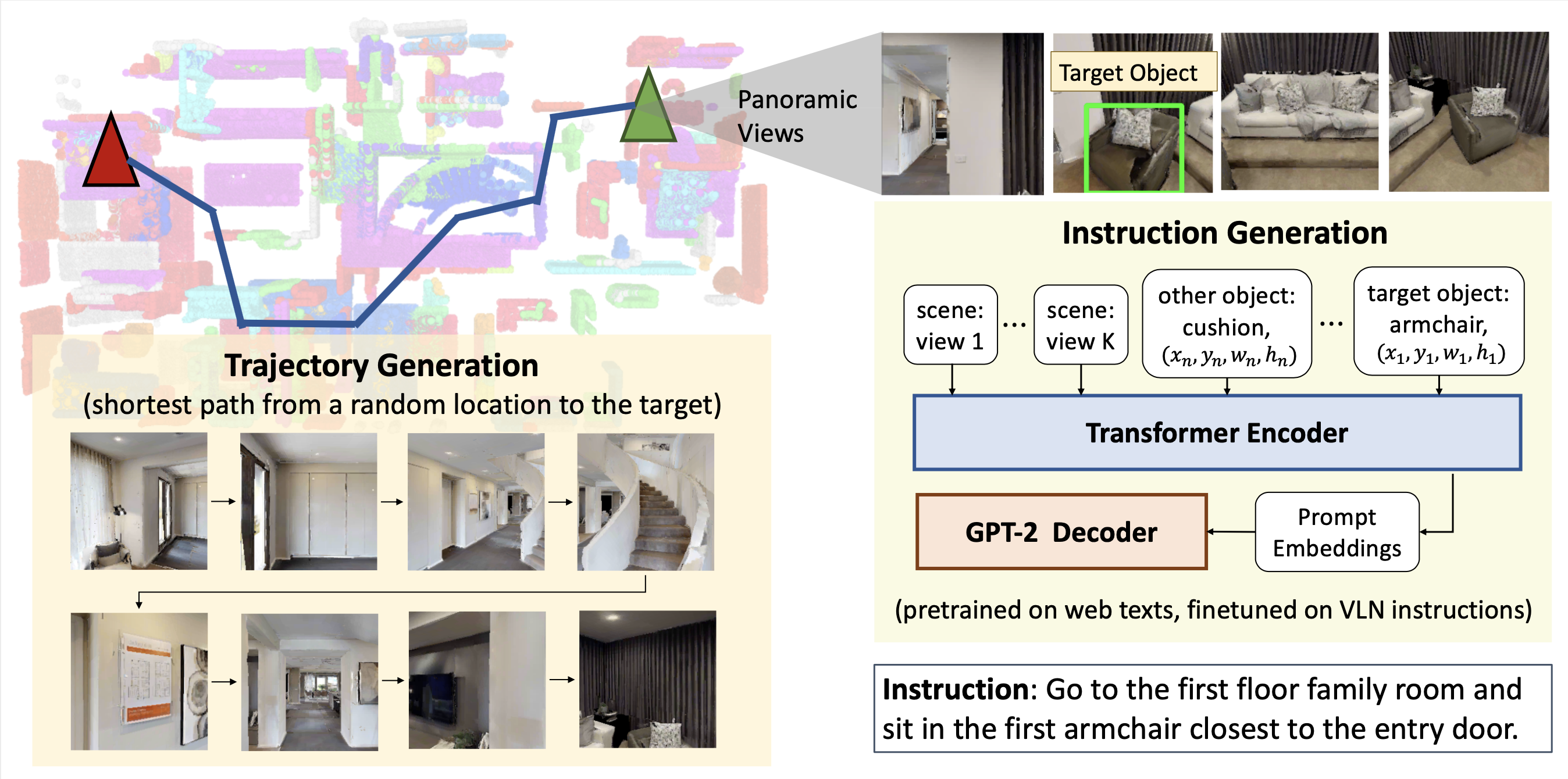}
\caption{To generate VLN training triplets, we first select a target object using the pseudo 3D object labels and then sample a trajectory using the navigation graph. Finally, we use a trained speaker model to generate high-level VLN instructions.}
\label{fig:data_generation_speaker}
\end{figure}

Based on the pseudo 3D object labels of the building, we generate object-trajectory-instruction triplets as shown in Fig.~\ref{fig:data_generation_speaker}.
For each 3D object, we obtain the corresponding goal locations in the graph where the object is visible and located within $d_o$ meters.
Then we randomly select a start node that is 4 to 9 steps away from the target location, and use the navigation graph to compute the shortest path as the expert trajectory.
The final panoramic image in the trajectory is used to generate high-level VLN instructions.

\subsubsection{Generating instructions with object prompts.}
Previous works~\cite{fried2018speaker,tan2019learning} train speaker models from scratch on VLN datasets to generate instructions from an image trajectory.
Due to the small amount of training examples in VLN datasets, such speaker models may suffer from inaccurate classification and are also unable to generate novel words beyond the training vocabulary. As a result, they even perform worse than template-based instructions on unseen images with diverse objects~\cite{guhur2021airbert}.
We alleviate these problems by (i) using object labels obtained from the 3D object labeling as prompts for the speaker model; and (ii) fine-tuning a pretrained large language model (GPT-2~\cite{radford2019gpt2}). 

To be specific, our speaker model consists of a transformer encoder that produces prompt embeddings and a language decoder that generates sentences conditioning on the prompts.
The encoder is fed with three types of tokens: target object token, other object tokens in the panorama, and view image tokens. The object token embedding encodes the object label, location, size, and visual representation, while the view image embedding is obtained from the visual representation and orientation embedding. A multi-layer transformer is adopted to learn relations of different tokens in order to generate more accurate referring expression for the target object.
We average the output embeddings for each type of tokens separately, resulting in three tokens (target object, other objects, and view image) as prompts to the decoder.
The decoder is initialized with pretrained GPT-2~\cite{radford2019gpt2}, a state-of-the-art language model. 
It takes the above prompts as the first three tokens and sequentially predicts words.
We finetune the speaker model end-to-end on the REVERIE dataset~\cite{qi2020reverie}. 

\begin{table}[t]
\centering
\tabcolsep=0.1cm
\caption{Statistics of training data on different VLN datasets. (gt.obj denotes whether the data relies on groundtruth object annotations)}
\label{tab:hm3d_autovln_stats}
\begin{tabular}{lcccccc} \toprule
dataset & gt.obj & \#env. & \#objs & \#instr & \#vocab & inst. length \\ \midrule
REVERIE \cite{qi2020reverie} & \checkmark & 60 & 12,029 & 10,466 & 1,140 & 18.64 \\
REVERIE-Speaker \cite{chen2022duet} & \checkmark & 64 & 12,062 & 19,636 & 399 & 15.20 \\
SOON \cite{zhu2021soon} & \checkmark & 34 & 4,358 & 2,780 & 735 & 44.09 \\ \midrule
HM3D-AutoVLN & $\times$ & 900 & 172,000 & 217,703 & 1,696 & 20.52\\ \bottomrule
\end{tabular}
\end{table}

\subsection{HM3D-AutoVLN: A Large-scale VLN Dataset}
\label{sec:hm3d_vln_dataset}

We employ the above procedure to process all unlabeled buildings in HM3D and generate a large-scale VLN dataset with high-level instructions, named HM3D-AutoVLN. 
Table~\ref{tab:hm3d_autovln_stats} compares HM3D-AutoVLN with existing VLN datasets that also use high-level instructions. 
REVERIE and SOON are manually annotated datasets and REVERIE-Speaker is an augmented dataset generated by a speaker model trained with groundtruth room and object annotations.
Our HM3D-AutoVLN is an order of magnitude larger than these datasets, in terms of environments, objects, and instructions.
Due to the pretrained language model, our dataset also features a diverse vocabulary\footnote{In Table~\ref{tab:hm3d_autovln_stats} vocab, we only count words that occur more than 5 times in the dataset.}  in the generated instructions.

%% file: method_model.tex
\section{VLN Model and Training}

In this section, we briefly introduce a state-of-the-art VLN model DUET~\cite{chen2022duet} and then describe its training procedure.

\subsection{DUET VLN Model}
The main idea of the DUal scalE graph Transformer (DUET)~\cite{chen2022duet} model is to build a topological map during navigation, which allows the model to make efficient long-term navigation plans over all navigable nodes in the graph instead of being limited to the neighboring nodes.
The model consists of two modules: topological mapping and global action planning.
The topological mapping module gradually adds newly observed locations to the map and updates node representations.
The global action planning module is a dual-scale graph transformer to predict a next location in the map or a stop action at each step.
The dual-scale architecture enables the model to jointly use fine-grained language grounding with fine-scale representations of the current location and global graph reasoning with coarse-scale representation of the map.
The DUET model is the winning entry in ICCV 2021 REVERIE and SOON navigation challenge. 
We use the publicly released code\footnote{\url{https://github.com/cshizhe/VLN-DUET}} in our experiments.

\subsection{Training}
\subsubsection{Stage I: Pretraining on HM3D-AutoVLN.}
We use three proxy tasks to pretrain DUET~\cite{chen2022duet}, including Masked Language Modeling (MLM), Single-step Action Prediction (SAP) and Object Grounding (OG).
For MLM, the inputs are a masked instruction and a full expert trajectory $(V_1, \cdots, V_T)$. The goal is to recover the masked words from the cross-modal visual representation.
For SAP, we ask the model to predict the next action given the instruction and its navigation history. We randomly select the navigation history as:
(i) $(V_1, \cdots, V_T)$ with target action \emph{STOP};
(ii) $(V_1, \cdots, V_t)$ where $1\leq t < T$ with target action $V_{t+1}$; and
(iii) a random node $V_i$ in the graph, and choose a target node $V_{i+1}$ with shortest overall distance from $V_i$ to $V_{i+1}$ and from $V_{i+1}$ to $V_T$ among all nodes in the constructed map.
Finally, for OG, the model is fed with an instruction and expert trajectory $(V_1, \cdots, V_T)$ to predict the ground-truth object in $V_T$.

\subsubsection{Stage II: Fine-tuning on downstream VLN datasets.}
We fine-tune a pretrained DUET on each downstream task by using the pseudo interactive demonstrator algorithm~\cite{chen2022duet} -- a special case of student forcing that addresses exposure bias.
We suggest two strategies for fine-tuning:
(i) fine-tuning on the downstream dataset only; or
(ii) balancing the size of HM3D-AutoVLN and the downstream dataset, and combining the two datasets for joint training.
The latter strategy is successful at preventing catastrophic forgetting of the previously learned policy as shown in our experiments in Table~\ref{tab:rvr_instr_vln}.

%% file: exprs.tex
\section{Experiments}

\subsection{Experiment Setup}

\subsubsection{Datasets.}
We evaluate models on two VLN tasks with high-level instructions, REVERIE~\cite{qi2020reverie} and SOON~\cite{zhu2021soon}.
Each dataset is divided into training, val seen, val unseen and a hidden test unseen split. 
The statistics of training splits are presented in Table~\ref{tab:hm3d_autovln_stats}.
The REVERIE dataset provides groundtruth object bounding boxes at each location and only requires an agent to select one of the bounding boxes. The shortest path from the agent's initial location to the target location is between 4 to 7 steps.
The SOON dataset instead does not give groundtruth bounding boxes, and asks an agent to directly predict the orientation of the object's center. The path lengths are also longer than REVERIE with 9.5 steps on average.
Due to the increased task difficulty and smaller training dataset size, it is more challenging to achieve high performance on SOON  than on REVERIE. 

\subsubsection{Evaluation metrics.}
We measure two types of metrics for the VLN task, navigation-only and remote object grounding.
The navigation-only metrics focus solely on whether an agent arrives at the target location. We use standard navigation metrics~\cite{anderson2018vision} including
\textbf{Success Rate (SR)}, the percentage of paths with the \emph{final} location near any target locations within 3 meters; 
\textbf{Oracle Success Rate (OSR)}, the percentage of paths with \emph{any} location close to target within 3 meters; and
\textbf{SR weighted by Path Length (SPL)} which multiplies SR with the ratio between the length of shortest path and the agent's predicted path.
As SPL takes into account navigation accuracy and efficiency, it is the primary metric in navigation.
The remote object grounding metrics~\cite{qi2020reverie} consider both navigation and object grounding performance.
The standard metrics are \textbf{Remote Grounding Success (RGS)} and \textbf{RGS weighted by Path Length (RGSPL)}.
RGS in REVERIE is defined as correctly selecting the object among groundtruth bounding boxes, while in SOON as predicting a center point that is inside of the groundtruth polygon.
RGSPL penalizes RGS by path length similar to SPL.
For all these metrics, higher is better.

\subsubsection{Implementation details.}
We adopt ViT-B/16~\cite{dosovitskiy2020image} pretrained on ImageNet to extract view and object features.
For SOON dataset, we use the Mask2Former trained on ADE20K~\cite{cheng2021mask2former} to extract object bounding boxes and transfer the prediction in the same way as in REVERIE.
We use the same hyper-parameters in modeling as the DUET model \cite{chen2022duet}.
All experiments were run on a single Nvidia RTX8000 GPU.
The best epoch is selected based on SPL on the val unseen split.

\subsection{Ablation Studies}
The main objective of our work is to explore how much VLN agents can benefit from large-scale synthesized dataset.
In this section, we carry out extensive ablations on the REVERIE dataset to study the effectiveness of our HM3D-AutoVLN dataset and key design choices.

\subsubsection{Contributions from HM3D-AutoVLN dataset.}
In Table~\ref{tab:rvr_instr_vln}, we compare the impact of VLN data generated from different sources for training DUET. 
Row 1 (R1) only relies on manually annotated instructions on 60 buildings in MP3D. Row 2 (R2) utilizes groundtruth room and object annotations to generate instructions for all objects in the 60 seen buildings. Such data is of high-quality and significantly improves VLN performance,~\eg~3\% on SPL.
In R3, we only use our generated HM3D-AutoVLN for pretraining, while still finetuning on the REVERIE train split.
Due to the visual diversity of the 900 additional buildings, it brings a larger boost than R2.
Compared to R1, the improvement of the SPL metric is 8.6\%, whereas other metrics like SR improve by 6.5\%.
This suggests that the pretraining on large-scale environments enables the model to learn efficient exploration, even without any additional manual annotations.
Moreover, when the model is fine-tuned jointly on the HM3D-AutoVLN and REVERIE datasets~(R4), the SR metric is further improved by 5\% over R3. This indicates that fine-tuning on the downstream dataset alone may suffer from forgetting and lead to worse generalization performance. 
Compared to the navigation metrics, the remote object grounding metrics see modest improvements.
We hypothesize that though our generated instructions often describe the object and scene accurately, they are not discriminative enough to refer to a specific object in the environment (\eg~unable to discriminate between multiple pillows placed on a sofa) or confuse predicting relationships between objects (\eg~second pillow from the left).
We see improving RGS and RGSPL as a promising future direction that would need to take into account relations between objects.

\begin{table}[t]
\centering
\tabcolsep=0.13cm
\caption{DUET performance on REVERIE (RVR) val unseen split using different training data. $^{\dagger}$ denotes manual object annotations are used to synthesize data}
\label{tab:rvr_instr_vln}
\begin{tabular}{lccccccc} \toprule
\multirow{2}{*}{} & \multicolumn{2}{c}{Training Data} & \multicolumn{3}{c}{Navigation} & \multicolumn{2}{c}{Grounding} \\
 & Pretrain & Finetune & OSR & SR & SPL & RGS & RGSPL \\ \midrule
R1 & RVR & RVR & 48.74 & 44.36 & 30.79 & 30.30 & 21.08 \\
R2 & RVR+Speaker$^{\dagger}$ & RVR & 51.07 & 46.98 & 33.73 & 32.15 & 23.03 \\
R3 & HM3D & RVR & 54.81 & 50.87 & 39.36 & 34.65 & \textbf{26.79} \\
R4 & HM3D & RVR+HM3D & \textbf{62.14} & \textbf{55.89} & \textbf{40.85} & \textbf{36.58} & 26.76 \\ \bottomrule
\end{tabular}
\end{table}

\begin{figure}[t]
    \centering
    \begin{subfigure}[b]{0.48\linewidth}
        \includegraphics[width=\linewidth]{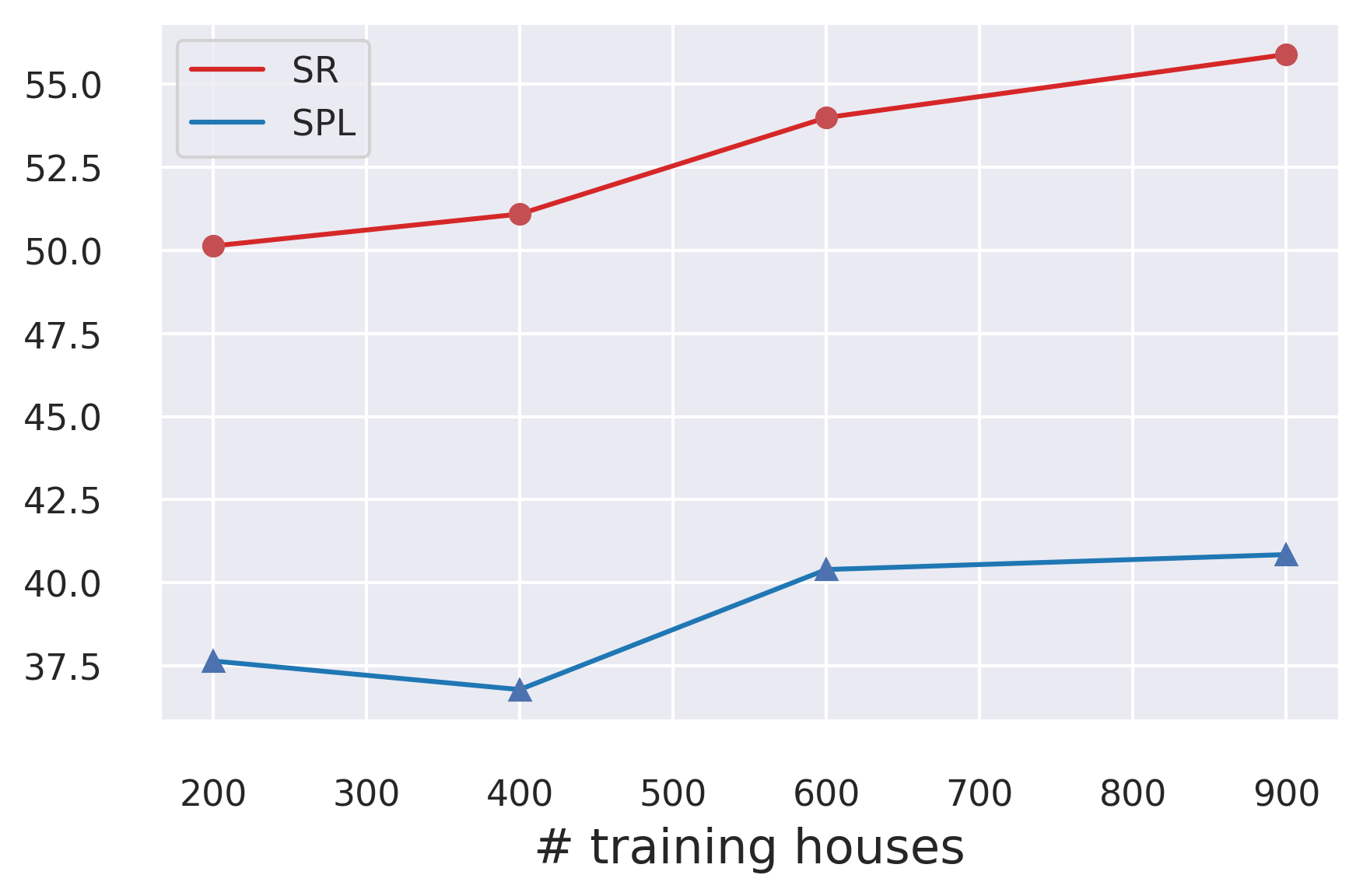}
        \caption{Navigation performance with respect to the number of training environments in HM3D-AutoVLN dataset.}
        \label{fig:rvr_vln_num_hm3d_environments}
    \end{subfigure}
    \hfill
    \begin{subfigure}[b]{0.48\linewidth}
        \centering
        \includegraphics[width=\linewidth]{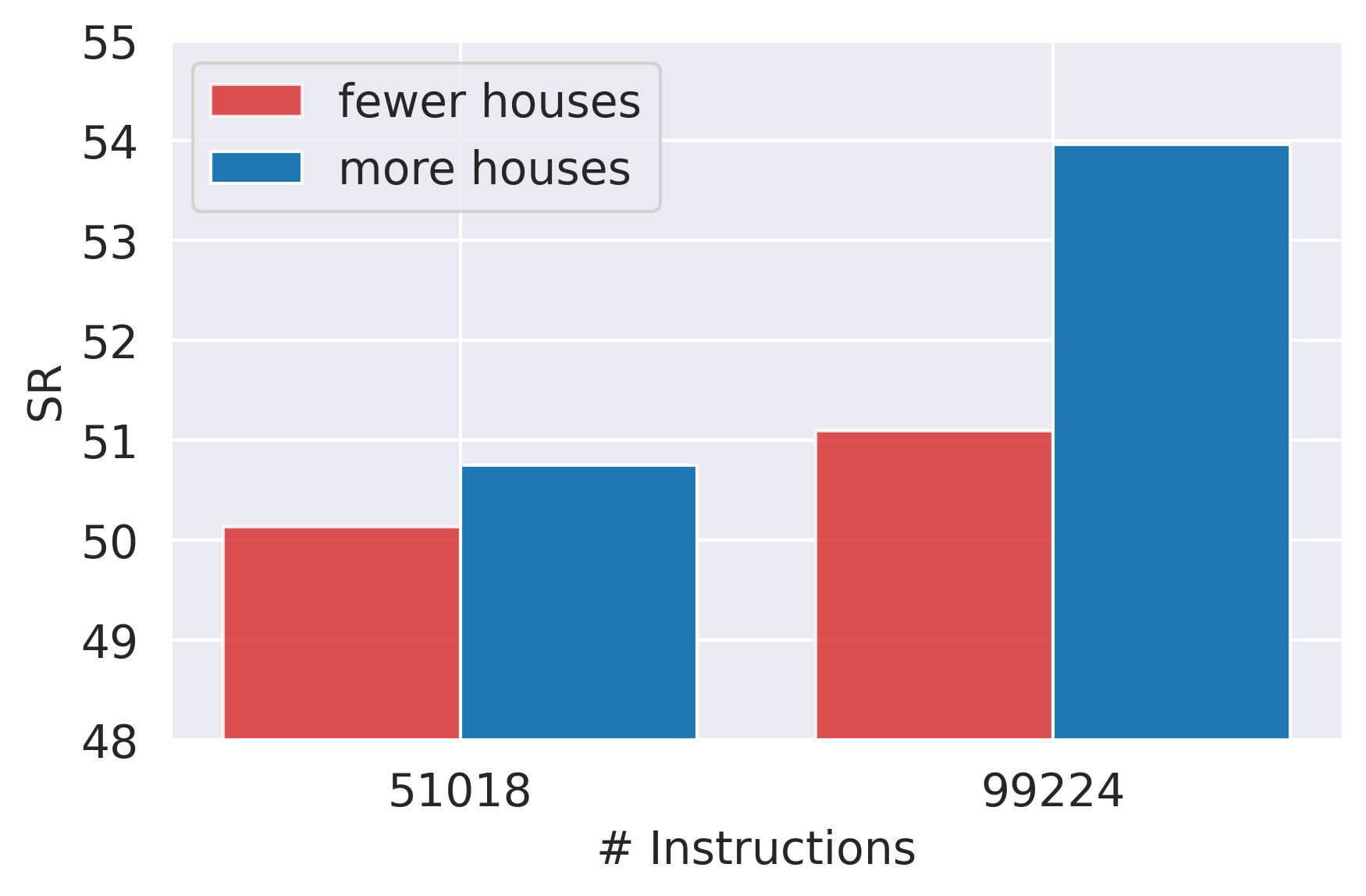}
        \caption{Given the same number of training examples, collecting data from more environments (blue) performs better than fewer environments.}
        \label{fig:rvr_environments_vs_instrs}
    \end{subfigure}
    \caption{Influence of the number of environments on DUET performance.}
    \label{fig:three graphs}
\end{figure}

\subsubsection{Impact of the number of training environments.}
Here, we evaluate the impact of the number of training environments. 
As pretraining on HM3D-AutoVLN mostly improves navigation performance, we mainly show navigation metrics SR and SPL. The full result table is provided in the supplementary material.
As shown in Figure~\ref{fig:rvr_vln_num_hm3d_environments}, more environments continuously improve the navigation performance.
We observe that even with the full 900 environments in HM3D, the gains are not saturated but increase gradually.
We also evaluate the impact of the number of environments given a fixed budget of VLN training examples.
Figure~\ref{fig:rvr_environments_vs_instrs} shows that using more environments improves the performance. 
On the \textbf{left}, the red bar uses 200 environments with 51,018 instructions, and the blue bar use the same amount of instructions but covers all 900 environments in HM3D. The \textbf{right} part is the same, but uses 400 environments for the red bar. 
The results suggest that it is beneficial to increase the number of environments rather than just increasing the trajectories for a limited number of environments.

\subsubsection{Finetuning with a small amount of supervised data.}
A large-scale automatically generated dataset improves pretraining and hence can reduce the supervised data requirement in downstream tasks.
We verify the effectiveness of HM3D-AutoVLN on a few-shot learning setting~\cite{guhur2021airbert} that compares the impact of finetuning on a variable number of REVERIE environments, see Table~\ref{tab:rvr_few_shot}.
Even without finetuning on any supervised instructions from REVERIE, our pretrained model achieves fair performance in comparison to a baseline that is only finetuned on all supervised data.
On the SPL metric, we outperform the baseline DUET model when finetuning on 10 environments (1/6 of the supervised data).
Moreover, finetuning with half the original data (30 environments) achieves significant boosts on all metrics compared to the baseline model.
A similar trend can be observed on the SOON dataset, see supplementary material.

\begin{table}[t]
\centering
\tabcolsep=0.13cm
\caption{DUET performance when using a fraction of the supervised data}
\label{tab:rvr_few_shot}
\begin{tabular}{cccccccc} \toprule
\multirow{2}{*}{\begin{tabular}[c]{@{}c@{}}HM3D\\ Pretrain\end{tabular}} & \multirow{2}{*}{\#environments} & \multirow{2}{*}{\#instructions} & \multicolumn{3}{c}{Navigation} & \multicolumn{2}{c}{Grounding} \\ 
 &  &  & OSR & SR & SPL & RGS & RGSPL \\ \midrule
$\times$ & 60 & 10,466 & 48.74 & 44.36 & 30.79 & 30.30 & 21.08  \\ \midrule
\checkmark & 0 & 0 & 43.08 & 36.81 & 25.28 & 20.82 & 13.74 \\
\checkmark & 1 & 449 & 50.78 & 42.12 & 29.55 & 25.02 & 17.26 \\
\checkmark & 10 & 1,404 & 50.47 & 43.79 & 33.61 & 26.30 & 20.24 \\
\checkmark & 30 & 5,244 & 60.81 & 53.71 & 39.26 & 34.42 & 25.11 \\
\checkmark & 60 & 10,466 & 62.14 & 55.89 & 40.85 & 36.58 & 26.76 \\ \bottomrule
\end{tabular}
\end{table}

\subsubsection{Comparing distance to objects.}
As mentioned in Sec.~\ref{sec:hm3d_vln_instr}, we select some of the visible objects that are close to the agent, within $d_o$ meters, to generate instructions.
In Table~\ref{tab:rvr_depth_obj_filter}, we compare the influence of different distances to the objects on the VLN performance.
Larger $d_o$ allows us to generate more instructions.
We can see that while including additional remote objects increases the number of instructions, it leads to a small drop in performance as the model struggles to identify objects that are small and far away.

\begin{minipage}[t]{\linewidth}
\begin{minipage}[t]{0.45\linewidth}
\makeatletter\def\@captype{table}
\tabcolsep=0.05cm
\scriptsize
\caption{DUET performance on instructions where the visible objects are at a different distance $d_o$ from the agent location}
\label{tab:rvr_depth_obj_filter}
\begin{tabular}{ccccccc} \toprule
\multirow{2}{*}{$d_o$} & \multirow{2}{*}{\#instrs} & \multicolumn{3}{c}{Navigation} & \multicolumn{2}{c}{Grounding} \\
&  & OSR & SR & SPL & RGS & RGSPL \\ \midrule
    2 & 217,703 &\textbf{62.14} & \textbf{55.89} & \textbf{40.85} & \textbf{36.58} & \textbf{26.76} \\
    3 & 396,401 & 57.37 & 53.25 & 40.38 & 34.28 & 25.65 \\ 
    $\infty$ & 544,606 & 59.98 & 53.37 & 38.03 & 35.70 & 25.50 \\ \bottomrule
\end{tabular}
\end{minipage}
\quad
\begin{minipage}[t]{0.48\linewidth}
\makeatletter\def\@captype{table}
\tabcolsep=0.05cm
\scriptsize
\centering
\caption{Comparison of different speaker models in terms of manual captioning evaluation and the followup navigation performance}
\label{tab:instr_eval}
\begin{tabular}{cccccc} \toprule
 & \multicolumn{3}{c}{Captioning} & \multicolumn{2}{c}{Navigation} \\
 & Room & Obj & Rel & SR & RGS \\ \midrule
Template & 0.13 & 0.76 & 0.05 & 52.20 & 32.75 \\
LSTM & 0.58 & 0.65 & 0.27 & 49.59 & 32.29 \\
GPT2 (Ours) & \textbf{0.73} & \textbf{0.78} & \textbf{0.35} & \textbf{55.89} & \textbf{36.58} \\ \bottomrule
\end{tabular}
\end{minipage}
\end{minipage}

\subsubsection{Evaluating quality of the HM3D-AutoVLN dataset.}
\begin{figure}[t]
    \scriptsize
    \begin{minipage}{.25\linewidth}
      \centering
                \includegraphics[width=\linewidth]{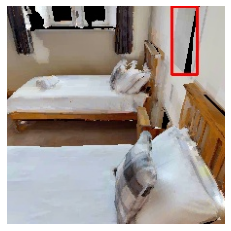}
                2D \texttt{window}, 3D \texttt{mirror}
    \end{minipage}%
    \begin{minipage}{.25\linewidth}
      \centering
                \includegraphics[width=\linewidth]{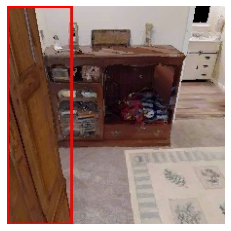}
                2D \texttt{door}, 3D \texttt{wardrobe}
    \end{minipage}
    \begin{minipage}{.25\linewidth}
      \centering
                \includegraphics[width=\linewidth]{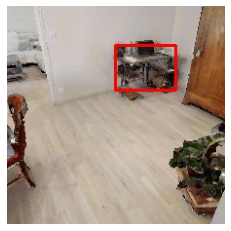}
                2D \texttt{table}, 3D \texttt{desk}
    \end{minipage}%
    \begin{minipage}{.3\linewidth}
      \centering
                \includegraphics[width=0.83\linewidth]{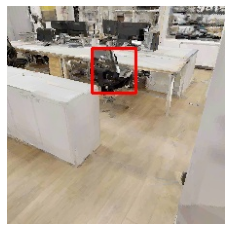}\\
                2D \texttt{chair}, 3D \texttt{swivel chair}
    \end{minipage}
        \caption{Qualitative examples of pseudo labelling}
        \label{fig:quality_pseudolabels}
\end{figure}

We validate each annotation procedure in our automatic dataset construction.
For \emph{navigation graph generation}, we measure whether the graph covers the whole building. Assuming each navigation node covers a circle with a radius of 2 meters, the graph achieves a high coverage rate of 93.4\% on average.
For \emph{3D object labeling}, we randomly select 300 bounding boxes and manually annotate semantic labels for them. We observe that 37.4\% of them are correctly predicted with 2D predictions, whereas 58.3\% of them were correctly predicted when applying cross-view consistency, showing an absolute improvement of 21\%.
Examples in Figure~\ref{fig:quality_pseudolabels} highlight the advantages of using cross-view consistency.
Due to the distorted view, it's reasonable that the 2D models wrongly predict the \texttt{mirror} and \texttt{wardrobe} to be a \texttt{window} and \texttt{door} respectively.
Cross-view consistency also helps to integrate scene context, swapping a \texttt{table} to a \texttt{desk} and a normal \texttt{chair} to a \texttt{swivel chair}.
For \emph{instruction generation}, we further compare our GPT2-based speaker model with template-based methods and a LSTM baseline~\cite{chen2022duet}. 
We manually evaluate 100 randomly selected instructions by measuring whether the instruction correctly mentions the target room, object class and object instance (with correct relations). We also measure the VLN performance using the generated instructions. Our model performs best on manual evaluation and on downstream VLN tasks as shown in Table~\ref{tab:instr_eval}.


\subsection{Comparison with State of the Art}
In Table~\ref{tab:reverie_sota_cmpr}, we compare with the state of the art on the REVERIE dataset.
To demonstrate the contributions of our automatically constructed dataset, we further train additional VLN agents with the augmentation of the HM3D-AutoVLN dataset, including EnvDrop~\cite{tan2019learning}, RecBert~\cite{hong2020recurrent} and HAMT~\cite{chen2021hamt}.
Our proposed dataset improves results of all methods and gives a particularly large boost to high-capacity models.
When pretraining DUET on HM3D-AutoVLN, the increase in performance over DUET with pretraining is significant for all metrics on both the val unseen and the test unseen splits. For example, the SPL measure increases by 7.1\% and 2.8\% on val unseen and test unseen splits of REVERIE.
Table~\ref{tab:soon_sota_cmpr} provides results for the SOON dataset.
Note that instructions from the SOON dataset are not used to train our speaker model (\cf~Sec.~\ref{sec:hm3d_vln_instr}) and are somewhat different from the REVERIE instructions.
Nevertheless, the large performance increase demonstrates cross-domain benefits of pretraining on an automatically collected large-scale dataset. Finally, we can also observe that the gain for object grounding is less significant, as discussed before this can be explained by the confusion between objects of the same categories and erroneous spatial relations.

\input{tables/reverie_sota}

\begin{table}[t]
\centering
\scriptsize
\tabcolsep=0.1cm
\caption{Comparison with the state-of-the-art methods on the SOON dataset}
\label{tab:soon_sota_cmpr}
\begin{tabular}{l|cccc|cccc} \toprule
\multirow{2}{*}{Methods} & \multicolumn{4}{c|}{Val Unseen} & \multicolumn{4}{c}{Test Unseen} \\
 & OSR & SR & SPL & RGSPL & OSR & SR & SPL & RGSPL \\ \midrule
GBE \cite{zhu2021soon} & 28.54 & 19.52 & 13.34 & 1.16 & 21.45 & 12.90 & 9.23 & 0.45 \\
DUET \cite{chen2022duet} & 50.91 & 36.28 & 22.58 & 3.75 & 43.00 & 33.44 & 21.42 & 4.17 \\  \midrule
DUET (+HM3D) & \textbf{53.19} & \textbf{41.00} & \textbf{30.69} & \textbf{4.06}  & \textbf{48.74} & \textbf{40.36} & \textbf{27.83} & \textbf{5.11} \\ \bottomrule
\end{tabular}
\end{table}

%% file: tables/reverie_sota.tex
\begin{table*}[t]
\scriptsize
\centering
\tabcolsep=0.09cm
\caption{Comparison with the state-of-the-art methods on REVERIE dataset}
\label{tab:reverie_sota_cmpr}
\begin{tabular}{l|ccccc|ccccc} \toprule
\multirow{3}{*}{Methods} & \multicolumn{5}{c|}{Val Unseen} & \multicolumn{5}{c}{Test Unseen} \\
& \multicolumn{3}{c}{Navigation} & \multicolumn{2}{c|}{Grounding} & \multicolumn{3}{c}{Navigation} & \multicolumn{2}{c}{Grounding}  \\ 
& OSR & SR & SPL & RGS & RGSPL & OSR & SR & SPL & RGS & RGSPL\\ \midrule
Human & - & - & - & - & - & 86.83 & 81.51 & 53.66 & 77.84 & 51.44 \\ \midrule
Seq2Seq \cite{anderson2018vision} & 8.07 & 4.20 & 2.84 & 2.16 & 1.63 & 6.88 & 3.99 & 3.09 & 2.00 & 1.58 \\
RCM \cite{wang2019reinforced} & 14.23 & 9.29 & 6.97 & 4.89 & 3.89 & 11.68 & 7.84 & 6.67 & 3.67 & 3.14 \\
SMNA \cite{ma2019self} & 11.28 & 8.15 & 6.44 & 4.54 & 3.61& 8.39 & 5.80 & 4.53 & 3.10 & 2.39 \\
\scriptsize{FAST-MAttNet} \cite{qi2020reverie} & 28.20 & 14.40 & 7.19 & 7.84 & 4.67 & 30.63 & 19.88 & 11.61 & 11.28 & 6.08 \\
SIA \cite{lin2021scene} & 44.67 & 31.53 & 16.28 & 22.41 & 11.56 & 44.56 & 30.80 & 14.85 & 19.02 & 9.20 \\
Airbert \cite{guhur2021airbert} & 34.51 & 27.89 & 21.88 & 18.23 & 14.18 & 34.20 & 30.28 & 23.61  & 16.83 & 13.28 \\ \midrule
EnvDrop~\cite{tan2019learning} &  26.3 & 23.0 & 19.9  & - & - & - & - & - & - & -\\
\quad \scriptsize{+HM3D-AutoVLN} &  29.3 & 25.2 & 20.6 & - & - & - & - & - & - & - \\ 
RecBERT (oscar) \cite{hong2020recurrent} & 27.66 & 25.53 & 21.06 & 14.20 & 12.00 & 26.67 & 24.62 & 19.48 & 12.65 & 10.00 \\ 
\quad \scriptsize{+HM3D-AutoVLN} & 33.23 & 29.20 & 23.12 & 18.12 & 14.18 & - & - & - & - & - \\ 
HAMT \cite{chen2021hamt} & 36.84 & 32.95 & 30.20 & 18.92 & 17.28 & 33.41 & 30.40 & 26.67 & 14.88 & 13.08 \\ 
\quad \scriptsize{+HM3D-AutoVLN}& 42.09 & 37.80 & 31.35 & 23.03 & 18.88 & - & - & - & - & - \\ 
DUET \cite{chen2022duet} & 51.07 & 46.98 & 33.73 & 32.15 & 23.03 & 56.91 & 52.51 & 36.06 & 31.88 & 22.06 \\  
\quad \scriptsize{+HM3D-AutoVLN} & \textbf{62.14} & \textbf{55.89} & \textbf{40.85} & \textbf{36.58} & \textbf{26.76} & \textbf{62.3} & \textbf{55.17} & \textbf{38.88} & \textbf{32.23} & \textbf{22.68} \\  \bottomrule

\end{tabular}
\end{table*}

%% file: conclusion.tex
\section{Conclusion}
This work addresses the lack of training data for VLN tasks. We propose to automatically generate pseudo 3D object labels and VLN instructions for a collection of large-scale unlabeled 3D environments.
Training on our new dataset HM3D-AutoVLN significantly improves VLN performance due to the large-scale pretraining or co-training. 
It also provides insights on the importance of dataset collection and the challenges inherent to leveraging unlabeled environments.
In particular, it shows that the diversity of navigation environments is more important than the number of training samples alone.

%% file: suppmat.tex
We first present additional automatically generated examples from our HM3D-AutoVLN dataset in Sec.~\ref{sec:dataset_examples}.
Then, we present further ablation analysis of our methods and also discuss results on the SOON dataset (equivalent analysis for the REVERIE dataset was in the main paper) in Sec.~\ref{sec:addition_ablations}.
Finally, some qualitative visualizations of predicted navigation results are presented in Sec.~\ref{sec:qualitative_results}.

\section{HM3D-AutoVLN Dataset}
\label{sec:dataset_examples}
\begin{figure}[p]
	\centering
	\includegraphics[width=1\linewidth]{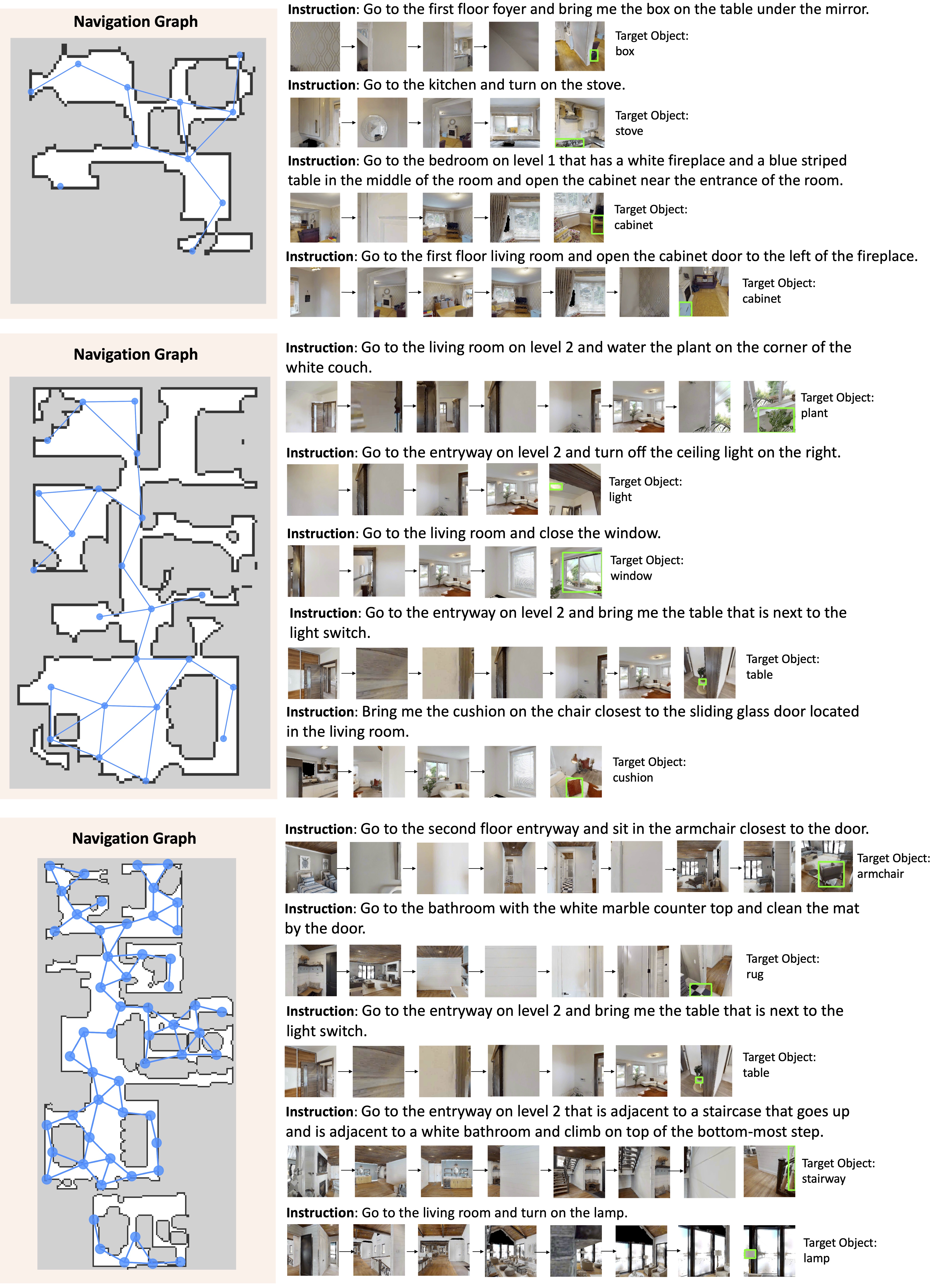}
	\caption{Examples of generated navigation graphs and object-trajectory-instruction triplets in HM3D-AutoVLN dataset.}
	\label{fig:hm3d_examples}
\end{figure}

In Fig.~\ref{fig:hm3d_examples}, we provide more examples of generated navigation graphs and object-trajectory-instruction triplets in the HM3D-AutoVLN dataset.
Each trajectory consists of a sequence of panoramas. We only visualize the oriented view image at each step in Fig.~\ref{fig:hm3d_examples} for simplicity.

\section{Additional Ablations}
\label{sec:addition_ablations}
\subsection{Comparison of Different Generated Instructions}
We compare our generated instructions from the proposed speaker model (Sec. 3.2 of the main paper) with two types of template-based instructions.
The first type of template instructions (Template.Obj) only describes the target object name, \eg, ``open the \underline{window}'', ``clean the \underline{mirror}''.
The second type (Template.Sent) uses sentence structures in REVERIE training split as templates~\cite{guhur2021airbert} and replaces the noun phrases in the template with detected objects in the scene, \eg, ``go to bedroom and clean the \underline{chandelier}''.
As these template instructions are less accurate and contain less details, we generate three instructions per target object to alleviate the impact of noisy instructions. 
As shown in Table~\ref{tab:instr_cmpr}, even with fewer number of instructions per object, the generated instructions from our speaker model outperform the two template-based instructions.

\begin{table}[h]
	\centering
	\tabcolsep=0.15cm
	\caption{DUET performance on the REVERIE val unseen split using different methods for generating instructions.}
	\label{tab:instr_cmpr}
	\begin{tabular}{cccccccc} \toprule
		\multirow{2}{*}{Method} & \multirow{2}{*}{\#Instrs} & \multirow{2}{*}{inst.len} & \multicolumn{3}{c}{Navigation} & \multicolumn{2}{c}{Grounding} \\
		&  & & OSR & SR & SPL & RGS & RGSPL \\ \midrule
		Template.Obj & 653,109 & 3.88 & 54.33 & 49.28 & 37.66 & 29.85 & 23.07 \\
		Template.Sent & 653,109 & 11.50 & 56.16 & 52.20 & 39.90 & 32.75 & 24.86 \\
		GPT2 (Ours) & 217,703 & 20.52 & \textbf{62.14} & \textbf{55.89} & \textbf{40.85} & \textbf{36.58} & \textbf{26.76}\\ \bottomrule
	\end{tabular} 
\end{table}

\begin{table}[h]
	\centering
	\tabcolsep=0.14cm
	\caption{DUET performance on the REVERIE val unseen split using different number of training environments from the HM3D-AutoVLN dataset.}
	\label{tab:num_pretrain_envs}
	\begin{tabular}{cccccc} \toprule
		\multirow{2}{*}{\#Envs} & \multicolumn{3}{c}{Navigation} & \multicolumn{2}{c}{Grounding} \\
		& OSR & SR & SPL & RGS & RGSPL \\ \midrule
		200 & 55.61 & 50.13 & 37.65 & 32.92 & 24.68 \\
		400 & 55.04 & 51.09 & 36.79 & 32.43 & 23.16 \\
		600 & 60.61 & 53.99 & 40.40 & 33.85 & 25.52 \\
		900 & \textbf{62.14} & \textbf{55.89} & \textbf{40.85} & \textbf{36.58} & \textbf{26.76} \\ \bottomrule
	\end{tabular}
\end{table}

\begin{table}[t]
	\centering
	\tabcolsep=0.14cm
	\caption{DUET performance on the REVERIE val unseen split using the same number of instructions but different number of training environments in HM3D-AutoVLN dataset (51,018 is the total number of instructions contained in the 200 training environments, and so is 99,224 in 400 training environments)}
	\label{tab:num_pretrain_envs_same_num_instrs}
	\begin{tabular}{ccccccc} \toprule
		\multirow{2}{*}{\#Instrs} & \multirow{2}{*}{\#Envs} & \multicolumn{3}{c}{Navigation} & \multicolumn{2}{c}{Grounding} \\
		&  & OSR & SR & SPL & RGS & RGSPL \\ \midrule
		\multirow{2}{*}{51,018} & 200 & 55.61 & 50.13 & 37.65 & 32.92 & 24.68 \\ 
		& 900 & 54.05 & 50.75 & 39.52 & 32.35 & 25.22 \\ \midrule
		\multirow{2}{*}{99,224} & 400 & 55.04 & 51.09 & 36.79 & 32.43 & 23.16 \\
		& 900 & 60.44 & 53.96 & 38.35 & 36.01 & 25.56 \\ \bottomrule
	\end{tabular}
\end{table}

\subsection{Influence of the Number of Training Environments}
Table~\ref{tab:num_pretrain_envs} and Table~\ref{tab:num_pretrain_envs_same_num_instrs} present the results for all evaluation metrics corresponding to the SR/SPL results in Fig.~4 of the main paper.
We can see that more training environments are beneficial for the performance on unseen environments in Table~\ref{tab:num_pretrain_envs} and that this result is reflected across all metrics.
When using a fixed number of instructions, it is better to collect training examples from more environments as shown in Table~\ref{tab:num_pretrain_envs_same_num_instrs}.

\subsection{Few-shot Results}
In Table~\ref{tab:soon_few_shot}, we show the performance of using different amounts of training data from the SOON dataset to fine-tune the DUET model pretrained on our HM3D-AutoVLN dataset.
This is comparable to Table~3 of the main paper that presented results on the REVERIE dataset.
To be noted, instructions in SOON dataset are not used to train our speaker model, and the style of instructions in HM3D-AutoVLN is quite different from the style in the SOON dataset. 
Despite the style differences, the pretraining allows our model to achieve better performance with a small amount of supervised data.

\begin{table}[t]
	\centering
	\tabcolsep=0.14cm
	\caption{DUET performance on SOON val unseen split using different number of SOON training environments in fine-tuning}
	\label{tab:soon_few_shot}
	\begin{tabular}{ccccccc} \toprule
		\multirow{2}{*}{Pretrain} & \multirow{2}{*}{\#Envs} & \multirow{2}{*}{\#Instrs} & \multicolumn{3}{c}{Navigation} & \multicolumn{1}{c}{Grounding} \\
		& &  & OSR & SR & SPL & RGSPL \\ \midrule
		$\times$ & 34 & 2,780 & 50.91 & 36.28 & 22.58 & 3.75 \\ \midrule
		\checkmark & 0 & 0 & 14.75 & 11.21 & 7.92 & 0.38 \\
		\checkmark & 1 & 67 & 37.32 & 25.22 & 17.97 & 2.17 \\
		\checkmark & 10 & 848 & 45.58 & 34.51 & 24.38 & 2.75 \\
		\checkmark & 34 & 2,780 & 53.19 & 41.00 & 30.69 & 4.06 \\ \bottomrule
	\end{tabular}
\end{table}

\section{Qualitative Results}
\label{sec:qualitative_results}

\begin{figure}[t]
    \centering
    \includegraphics[width=0.8\linewidth]{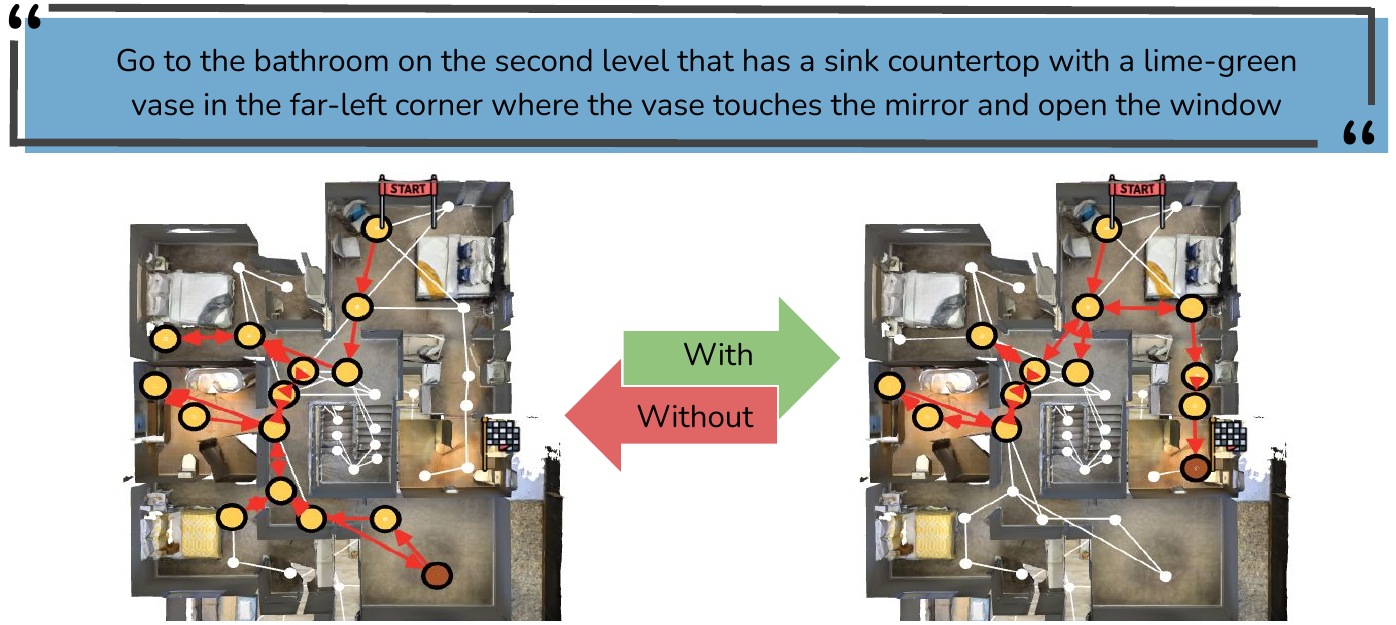}
    \includegraphics[width=0.8\linewidth]{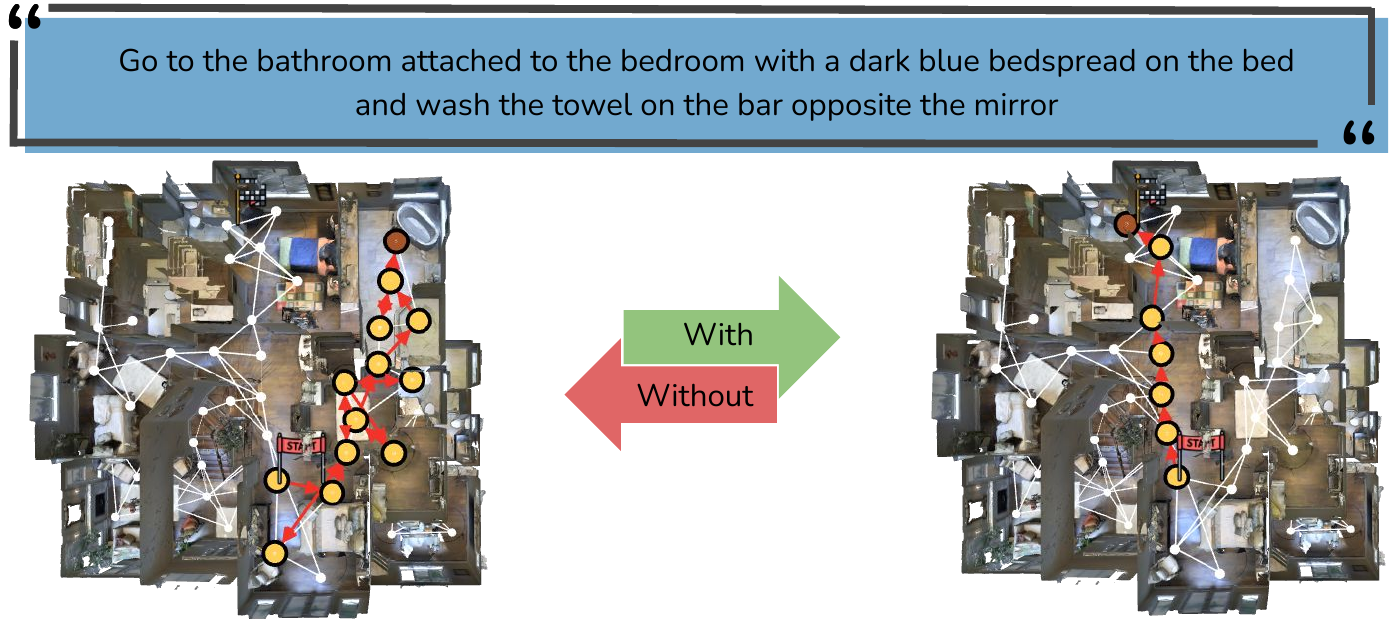}
    \includegraphics[width=0.8\linewidth]{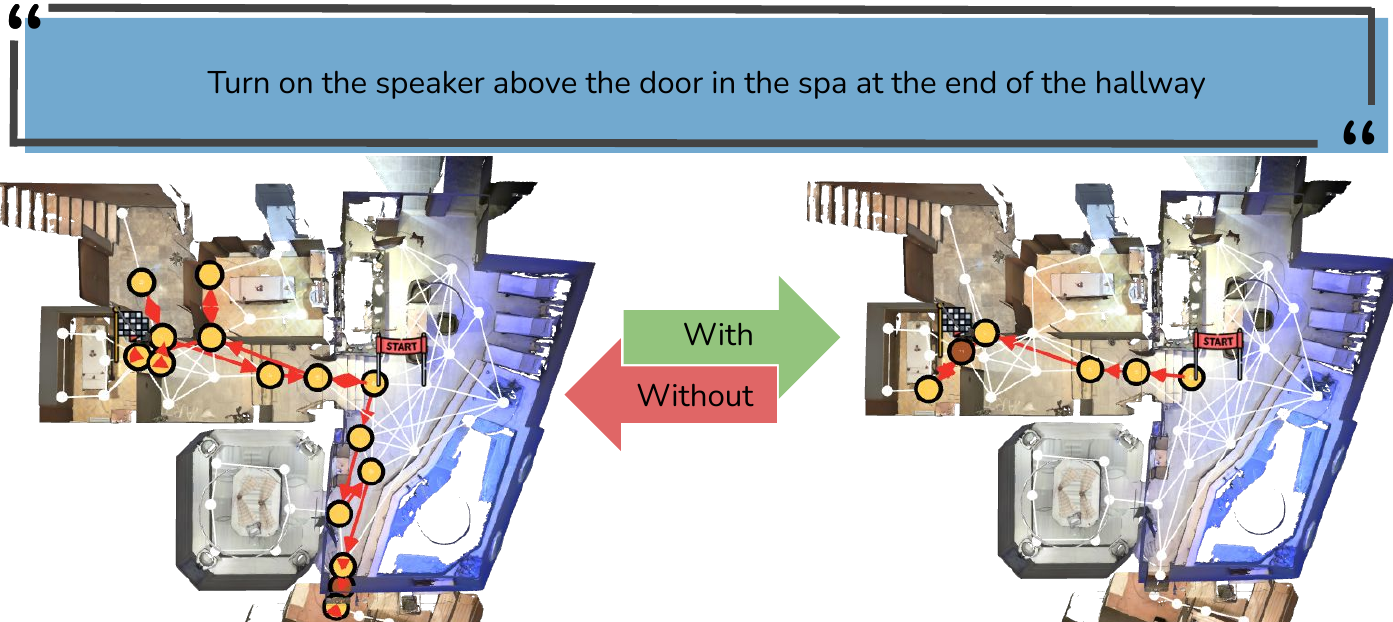}
    \includegraphics[width=0.8\linewidth]{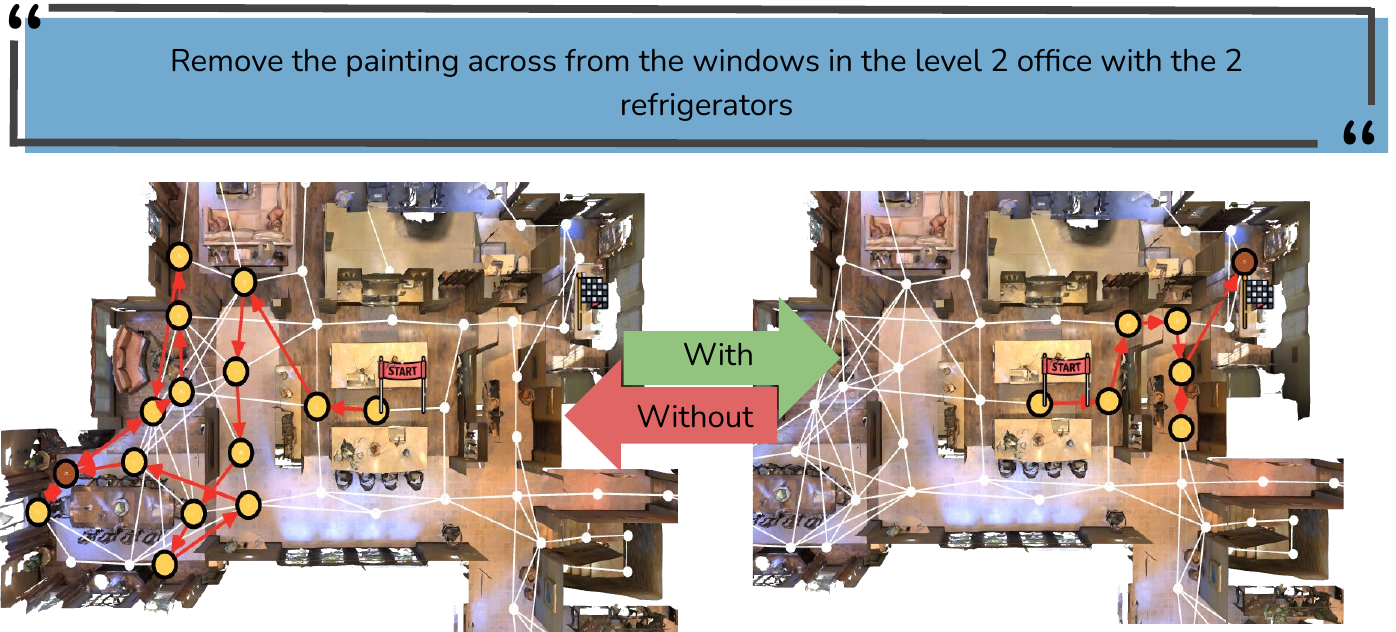}
    \caption{Examples where pretraining with HM3D-AutoVLN improved the performance of DUET on REVERIE val unseen split. The checkered flag and maroon node denote the target and predicted destinations respectively.}
    \label{fig:vln_success_cases}
\end{figure}

\begin{figure}[t]
    \centering
    \includegraphics[width=0.8\linewidth]{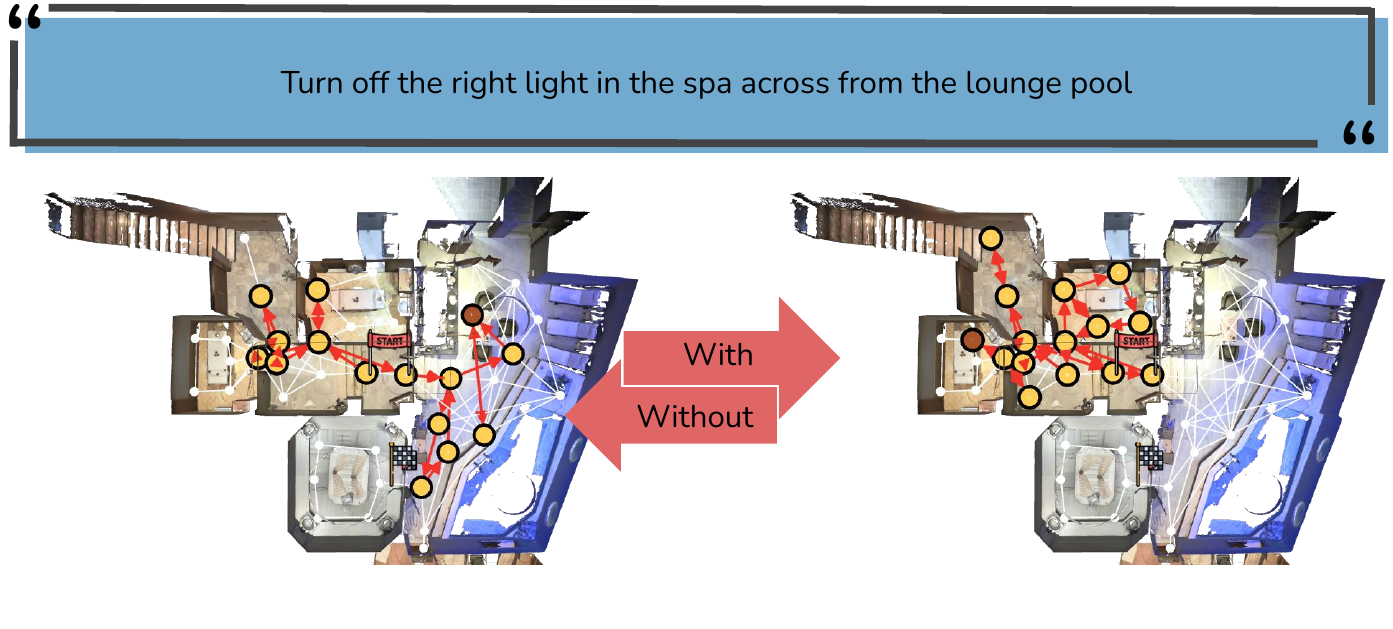}
    \includegraphics[width=0.8\linewidth]{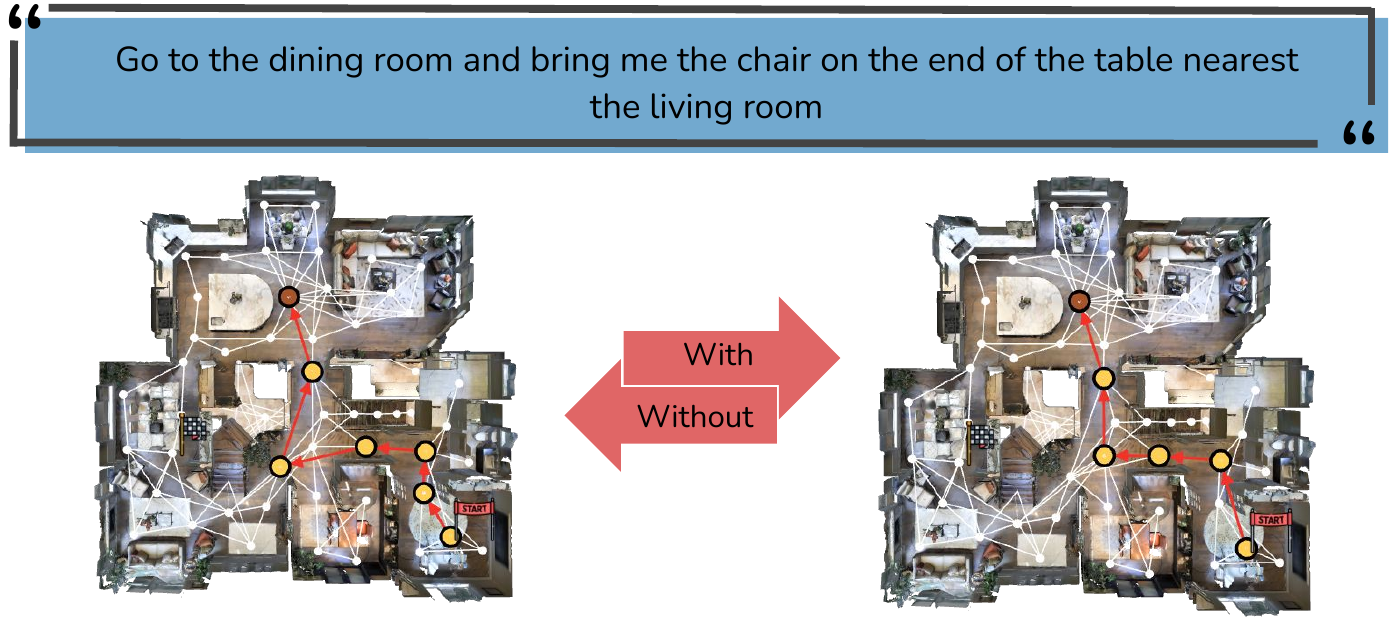}
    \includegraphics[width=0.8\linewidth]{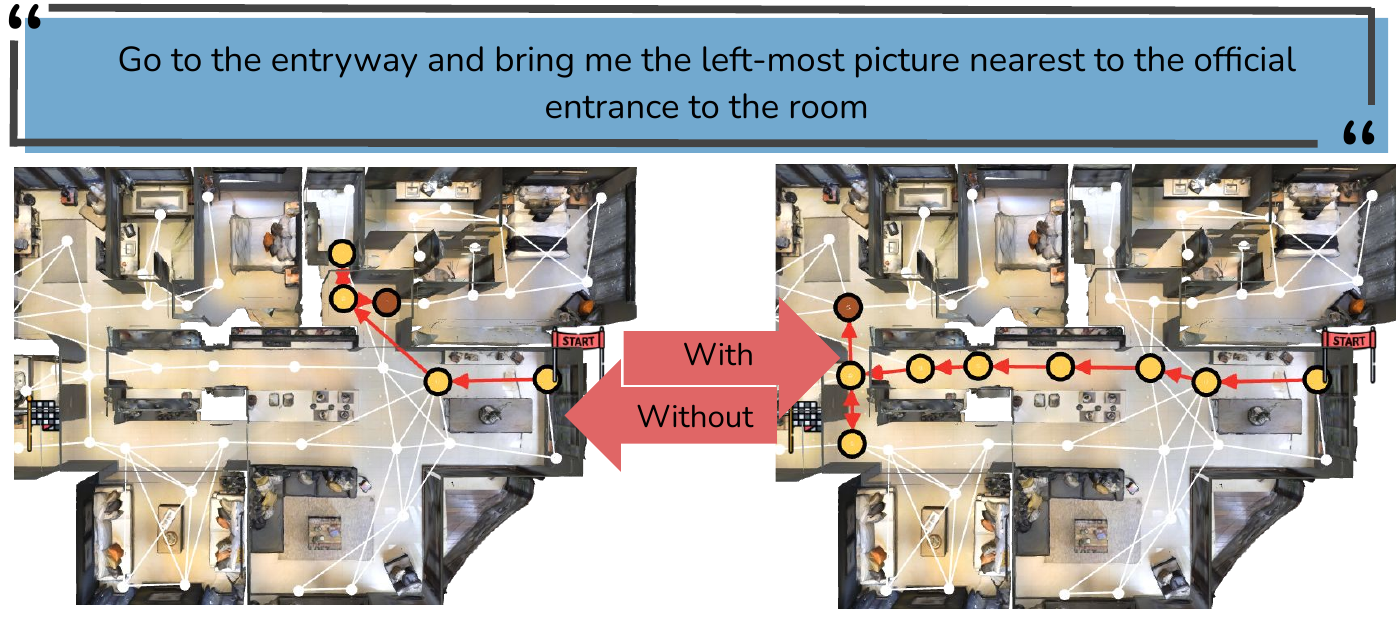}
    \includegraphics[width=0.8\linewidth]{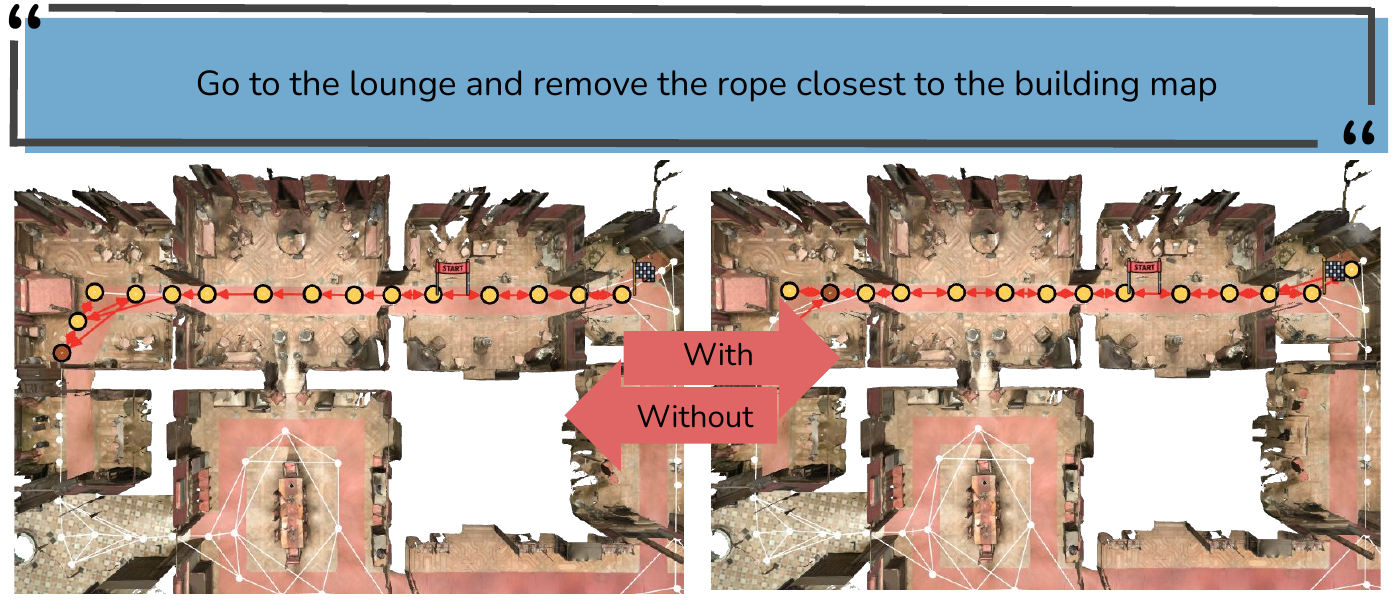}
    \caption{Examples where DUET models with and without pretraining on HM3D-AutoVLN both fail on REVERIE val unseen split. The checkered flag and maroon node denote the target and predicted destinations respectively.}
    \label{fig:vln_failure_cases}
\end{figure}

Fig.~\ref{fig:vln_success_cases} compares some predicted trajectories from DUET models with and without pretraining on our HM3D-AutoVLN dataset.
We can see that the pretraining allows the navigation model to explore the unseen environment more efficiently.
Fig.~\ref{fig:vln_failure_cases} further presents some failure cases. The main problem is to correctly decide the stop location aligned to the fine-grained instruction. For example, in the last two examples in Fig.~\ref{fig:vln_failure_cases}, the agent has navigated close to the target location but failed to stop and continued the exploration.

